\documentclass{ecai} 

\usepackage{latexsym}
\usepackage{amssymb}
\usepackage{amsmath}
\usepackage{amsthm}
\usepackage{booktabs}
\usepackage{enumitem}
\usepackage{graphicx}
\usepackage{color}
\usepackage{pgfplots}
\pgfplotsset{compat=1.18}
\usepackage{fix-cm}

\usepackage{cite}
\usepackage{amsfonts}
\usepackage{algorithmic}
\usepackage{algorithm}

\usepackage[citecolor=black]{hyperref}

\usepackage{mathtools}
\usepackage{dsfont}
\usepackage{subfigure}

\usepackage{multirow}
\usepackage{xspace}

\newcommand{\sysname}{\texttt{BoBa}\xspace}

\usepackage{bibentry}
\usepackage{newfloat}
\usepackage{listings}

\usepackage{tikz}
\usepackage{collcell}
\usepackage{etoolbox}
\newtoggle{inTableHeader}
\toggletrue{inTableHeader}

\newcommand*{\StartTableHeader}{\global\toggletrue{inTableHeader}}%
\newcommand*{\EndTableHeader}{\global\togglefalse{inTableHeader}}%

\let\OldTabular\tabular%
\let\OldEndTabular\endtabular%
\renewenvironment{tabular}{\StartTableHeader\OldTabular}{\OldEndTabular\StartTableHeader}%

\newcommand*{\MinNumber}{0}%
\newcommand*{\MidNumber}{0.08} %
\newcommand*{\MaxNumber}{1.0}%

\definecolor{lightred}{rgb}{1.0, 0.8, 0.8}   
\definecolor{lightgreen}{rgb}{0.0, 1.0, 0.0} 

\newcommand{\ApplyGradient}[1]{%
  \iftoggle{inTableHeader}{#1}{%
    \ifdim #1 pt > \MidNumber pt
      \pgfmathsetmacro{\PercentColor}{max(min(100.0*(#1 - \MidNumber)/(\MaxNumber-\MidNumber),100.0),0.00)}%
      \tikz[baseline=(X.base)] \node[fill=lightred!\PercentColor!yellow, inner sep=0pt, outer sep=0pt, text height=1.5ex, text depth=0.5ex, minimum width=0.8cm] (X) {\makebox[0.8cm][c]{#1}};%
    \else
      \pgfmathsetmacro{\PercentColor}{max(min(100.0*(\MidNumber - #1)/(\MidNumber-\MinNumber),100.0),0.00)}%
      \tikz[baseline=(X.base)] \node[fill=lightgreen!\PercentColor!yellow, inner sep=0pt, outer sep=0pt, text height=1.5ex, text depth=0.5ex, minimum width=0.8cm] (X) {\makebox[0.8cm][c]{#1}};%
    \fi
  }
}

\newcolumntype{R}{>{\collectcell\ApplyGradient}c<{\endcollectcell}}

\newcommand*{\MinNumberr}{0.983}%
\newcommand*{\MidNumberr}{0.99} %
\newcommand*{\MaxNumberr}{0.995}%

\newcommand{\ApplyGradientt}[1]{%
  \iftoggle{inTableHeader}{#1}{%
    \ifdim #1 pt > \MidNumber pt
      \pgfmathsetmacro{\PercentColor}{max(min(100.0*(#1 - \MidNumberr)/(\MaxNumberr-\MidNumberr),100.0),0.00)}%
      \tikz[baseline=(X.base)] \node[fill=lightgreen!\PercentColor!yellow, inner sep=0pt, outer sep=0pt, text height=1.5ex, text depth=0.5ex, minimum width=0.8cm] (X) {\makebox[0.8cm][c]{#1}};%
    \else
      \pgfmathsetmacro{\PercentColor}{max(min(100.0*(\MidNumberr - #1)/(\MidNumberr-\MinNumberr),100.0),0.00)}%
      \tikz[baseline=(X.base)] \node[fill=lightred!\PercentColor!yellow, inner sep=0pt, outer sep=0pt, text height=1.5ex, text depth=0.5ex, minimum width=0.8cm] (X) {\makebox[0.8cm][c]{#1}};%
    \fi
  }
}

\newcolumntype{M}{>{\collectcell\ApplyGradientt}c<{\endcollectcell}}

\newtheorem{theorem}{Theorem}

\newcommand{\BibTeX}{B\kern-.05em{\sc i\kern-.025em b}\kern-.08em\TeX}

\begin{document}


\begin{frontmatter}


\paperid{4818} 


\title{BoBa: Boosting Backdoor Detection through Data\\
Distribution Inference in Federated Learning}


\author[A]{\fnms{Zhengyuan}~\snm{Jiang}\footnote{Equal contribution.}}
\author[B]{\fnms{Xingyu}~\snm{Lyu}\footnotemark}
\author[C]{\fnms{Shanghao}~\snm{Shi}} 
\author[D]{\fnms{Yang}~\snm{Xiao}} 
\author[B]{\fnms{Yimin}~\snm{Chen}}
\author[C]{\fnms{Y.~Thomas}~\snm{Hou}} 
\author[C]{\fnms{Wenjing}~\snm{Lou}} 
\author[A]{\fnms{Ning}~\snm{Wang}\thanks{Corresponding Author. Email: ningw@usf.edu.}}

\address[A]{University of South Florida, USA}
\address[B]{University of Massachusetts, Lowell, USA}
\address[C]{Virginia Polytechnic Institute and State University, USA}
\address[D]{University of Kentucky, USA}

\begin{abstract}
Federated learning, while being a promising approach for collaborative model training, is susceptible to backdoor attacks due to its decentralized nature. Backdoor attacks have shown remarkable stealthiness, as they compromise model predictions only when inputs contain specific triggers.
As a countermeasure, anomaly detection is widely used to filter out backdoor attacks in FL. However, the non-independent and identically distributed (non-IID) data distribution nature of FL clients presents substantial challenges in backdoor attack detection, as the data variety introduces variance among benign models, making them indistinguishable from malicious ones. 

In this work, we propose a novel distribution-aware backdoor detection mechanism, BoBa, to address this problem. To differentiate outliers arising from data variety versus backdoor attacks, we propose to break down the problem into two steps: clustering clients utilizing their data distribution, and followed by a voting-based detection. We propose a novel data distribution inference mechanism for accurate data distribution estimation. To improve detection robustness, we introduce an overlapping clustering method, where each client is associated with multiple clusters, ensuring that the trustworthiness of a model update is assessed collectively by multiple clusters rather than a single cluster. Through extensive evaluations, we demonstrate that BoBa can reduce the attack success rate to lower than 0.001 while maintaining high main task accuracy across various attack strategies and experimental settings.
\end{abstract}

\end{frontmatter}

\section{Introduction}

Federated learning (FL) is gaining popularity for its privacy advantage for users over traditional centralized machine learning \citep{konevcny2016federated1, mcmahan2017communication, kairouz2021advances}, with applications including Android Gboard for next-word prediction \citep{hard2018federated} and WeBank's credit risk control \citep{wang2019adaptive}. While the distributed nature preserves the data privacy of individual clients, it creates opportunities for malicious clients to \emph{backdoor} the global FL model \citep{bhagoji2019analyzing, xie2019dba,wang2020attack}. 
Backdoor attacks aim to direct the model to make a desired prediction on targeted inputs while maintaining a relatively high accuracy on non-targeted inputs.

\begin{figure}[tbp]
    \centering
    \includegraphics[width=0.9\linewidth]{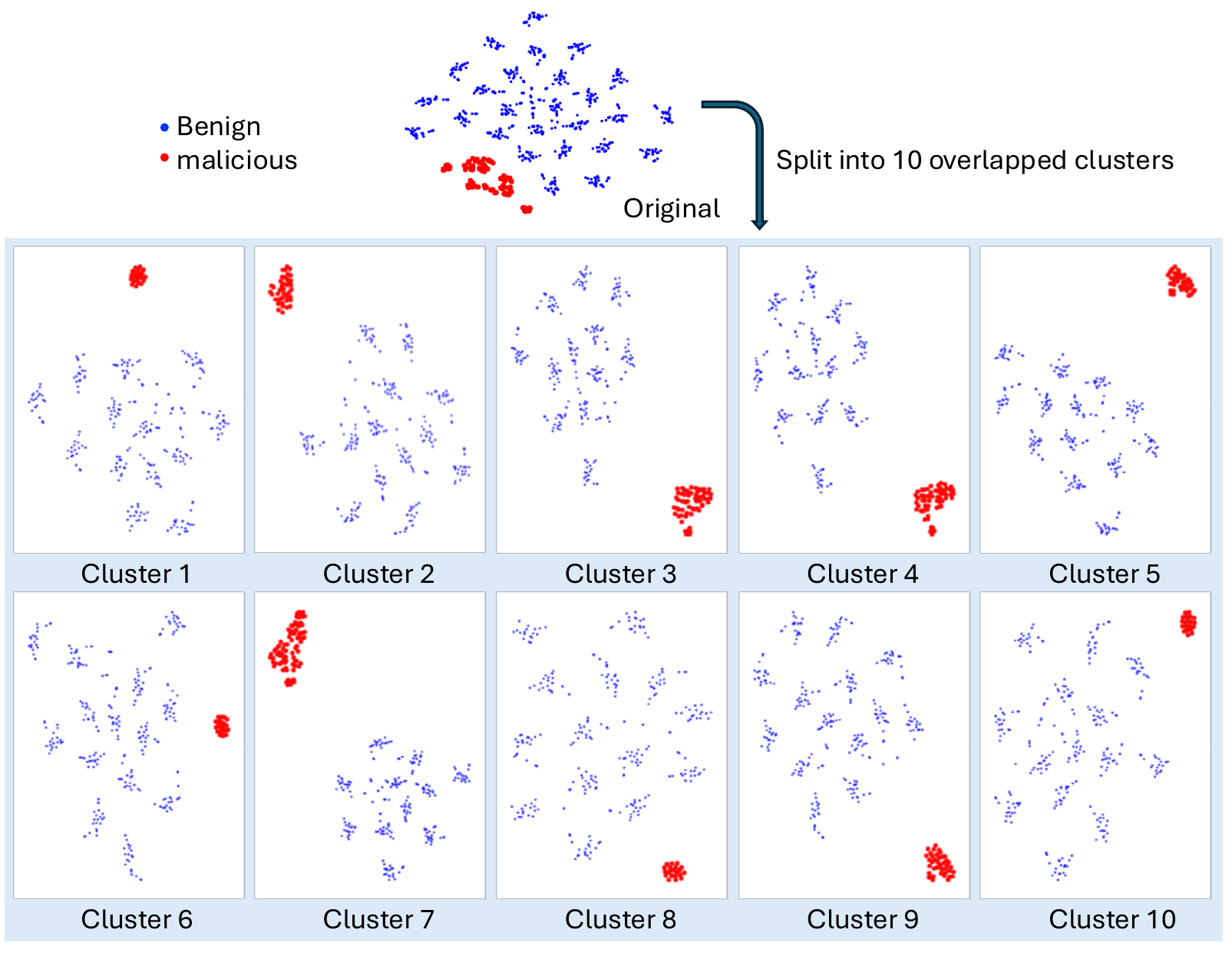}
    \vspace{-5pt}
    \caption{TSNE of client models in non-iid data scenario without clustering and with clustering (BoBa). }
    \label{fig:non-iid-illustration}
    \vspace{15pt}
\end{figure}

\emph{\textbf{The Non-IID Data Challenge.~}} The non-IID data in real-world FL systems further complicates the backdoor detection task. Most current defense mechanisms designed for IID data heavily rely on the assumption that a distinction exists between malicious model updates and benign model updates in a certain feature space \citep{shen2016auror, fung2018mitigating,li2020learning}. Such solutions analyze and extract the most distinguishable features, e.g., the angular distance of gradient updates \citep{fung2018mitigating}, low-dimensional embeddings of model parameters \citep{li2020learning}, model accuracy on generated data \citep{zhao2019pdgan}, etc. However, as illustrated in the top left figure (labeled as original) of Figure \ref{fig:non-iid-illustration}, we observe that malicious models are not differentiable from benign models in non-IID scenarios, which is similarly found in \citep{rieger2022deepsight,awan2021contra, briggs2020federated}. After applying \sysname, clients are divided into multiple clusters. Within each cluster, malicious clients are obviously distinct from benign models.

Clustering method has been explored by multiple existing works. Ghosh et al. \citep{ghosh2019robust} propose clustering clients based on their empirical risk value and then using trimmed mean \citep{yin2018byzantine} within each cluster to filter out possible Byzantine nodes. Rieger et al. \citep{rieger2022deepsight} suggest employing an ensemble of clustering algorithms to effectively cluster model updates with similar training data. \citep{rieger2024crowdguard} proposes a stacked clustering method, that selects the most representative voting vector from all submissions. Existing works utilize model weights for clustering without considering the data distribution differences. To cleanse the received model updates, the server will inevitably remove novel benign clusters from the final aggregation. It is a sacrifice of efficiency/utility for security. Our method aims to better balance utility and security by reducing the probability of removing novel benign clients.

To achieve the goal, we propose \sysname, a \textbf{Bo}oster for \textbf{Ba}ckdoor attack detection. \sysname introduces a novel clustering strategy based on a data distribution inference module, dubbed DDIG---\emph{Data Distribution Inference from Gradients}. DDIG can accurately estimate the relationship between gradients and label distribution theoretically and empirically. Based on the label distribution, we cluster the clients into multiple groups. In each cluster, clients hold similar data distributions. We successfully transformed the challenging backdoor detection in a non-IID scenario into an IID case. Further, we identify that a naive clustering method is vulnerable to malicious collisions where multiple attackers obtain the same distribution and dominate one cluster. We propose two constraints, \emph{uniform cluster inclusion} and \emph{balanced cluster}, to decrease the impact of collusion attack. Our system is a booster as we can attach the proposed clustering method to various traditional backdoor detection mechanisms designed for IID scenarios.

In summary, we make the following contributions:
\begin{itemize}
    \item We propose \sysname to address the challenging problem of backdoor detection for FL systems in non-IID data scenarios. To distinguish backdoored models from benign ones in non-IID data scenarios, we've developed a two-step detection method involving client clustering by data distribution and subsequent detection of malicious model updates within each cluster.
    \item We find out that the lack of knowledge of client data distribution is a key reason for the low accuracy of backdoor detection in non-IID scenarios. We propose a novel data distribution inference model--DDIG, to address this challenge, which has significantly improved the detection performance.
    \item We propose an overlapping clustering method to improve the robustness of backdoor attack detection. We model clustering as an optimization problem and address these challenges by incorporating balanced cluster and uniform cluster inclusion into the objective function. 
    \item Our evaluation showcases the superiority of \sysname over other baseline methods, as it consistently achieves a lower attack success rate (ASR) across various attack strategies and non-IID levels on multiple datasets.    
\end{itemize}

\section{Background and Related Work}
\subsection{Federated Learning}
\label{sec:system-model}
In FL systems, there are two entities, one parameter server (PS), and $n$ clients (we define $[n]\coloneqq\{1,2,...,n\}$). Each client manages a local dataset $\mathcal{D}_i$ following non-identical distributions. We use $m$ to represent the total number of data classes. We assume that each client manages a local model, and the model parameter of client $i$ is denoted by $\mathbf{\theta}_i\in \mathcal{W}\subseteq \mathbb{R}^d$, wherein $\mathcal{W}$ is the parameter space and $d$ is the presumed model dimensionality. The global model parameter is denoted by $\Theta\in \mathcal{W}$. We denote the model update from client $i$ as $\delta_i=\mathbf{\theta}_i - {\Theta}$. PS acts as the model distributor and aggregator on the cloud side. 
The entropy-based loss value for a local model is $\mathcal{L}(\theta_i, \mathcal{D}_i)$. The total loss function $\mathcal{F}(\Theta)$ is calculated using the loss of the $k$ selected clients, which can be formulated as 
\begin{equation}
    \mathcal{F}(\Theta) \coloneqq \sum^{k}_{i=1} \mathcal{L}(\theta_i, \mathcal{D}_i)
\end{equation}
The goal of the FL system is to jointly minimize the loss $\mathcal{F}(\Theta)$ by optimizing the global model parameters $\Theta$.

\subsection{Poisoning Attacks in FL}
In FL, since attackers are able to manipulate both data and models, a poisoning attack in FL is referred to as a model poisoning attack (MPA) \citep{chen2017distributed, so2020byzantine, guerraoui2018hidden, fang2020local, bhagoji2019analyzing}. Attackers carefully manipulate model parameters, with the aim of gradually degrading the FL model efficacy without being detected. Based on the goals of attackers, MPA can be categorized into two classes: untargeted attacks that aim at increasing the overall prediction error \citep{fang2020local} and backdoor attack/targeted attacks that manipulate the prediction on targeted inputs \citep{bhagoji2019analyzing,bagdasaryan2020backdoor}. We focus on the far more stealthy backdoor attacks in this work. Backdoor attacks have advanced their performance by enhancing {stealthiness} and {sophistication} \citep{xie2019dba,sun2019can, bhagoji2019analyzing, bagdasaryan2020backdoor}. 

\textbf{BadNet}\citep{gu2017badnets} serves as a foundational attack, illustrating straightforward trigger injection without model‐level manipulations. \textbf{Alternate}\citep{bhagoji2019analyzing} alternates between optimizing classification accuracy and minimizing deviation from benign models to achieve stealthy yet effective backdoor insertion. \textbf{DBA}\citep{xie2019dba} decomposes and distributes global triggers across clients to enhance persistence and concealment. \textbf{Sybil} \citep{fung2018mitigating} leverages multiple colluding malicious clients sharing the same crafted update to amplify the backdoor’s influence. \textbf{Neurotoxin}~\citep{zhang2022neurotoxin} exploits parameters with minimal updates during training to maintain backdoor persistence. \textbf{IBA}~\citep{nguyen2024iba} advances further by embedding stealthy and irreversible backdoors, offering strong resistance to existing defenses. These attacks span simple injection, optimization‐based stealth, distributed triggering, multi‐client collusion, minimal‐update persistence, and irreversible embedding, providing a rigorous testbed for assessing defense mechanisms in adversarial federated learning environments.

\subsection{Defense against Poisoning Attack} 
In the literature, the initial defenses, e.g., \textit{Krum} \citep{blanchard2017machine}, \textit{Median}, \textit{Trim} \citep{yin2018byzantine}, and \textit{Bulyan} \citep{guerraoui2018hidden}, typically leverage outlier-robust measures to compute the center of updates to filter out Byzantine updates. These mechanisms provide provable resilience against poisoning attacks to some extent. \citep{fung2018mitigating} utilizes angular distance of gradient updates for detection. Low-dimensional embeddings extracted from model parameter update are utilized in \citep{li2020learning}. \citep{rieger2024crowdguard, wang2022flare, zhao2019pdgan} explicitly detects and filters abnormal model updates using model representations/accuracy.

Later, trust-based approaches \citep{cao2021fltrust, awan2021contra,ali2024adversarially} and clustering-based approaches \citep{MESAS, fereidooni2023freqfed, rieger2024crowdguard} are proposed to further enhance the defense landscape.
\textbf{FLTrust} \citep{cao2021fltrust} trains a server model using an auxiliary dataset at PS in each iteration. PS bootstraps a trust score for each client based on its directional deviation from the server model update. 
\textbf{CONTRA} \citep{awan2021contra} detects malicious clients by evaluating the alignment of any two clients' gradients. Malicious client pairs have larger alignments than other pairs. CONTRA implements a cosine similarity-based measure to determine the credibility of local model parameters.
\textbf{MESAS}~\citep{MESAS} used a multi-faceted detection strategy, statistical tests, clustering, pruning, and filtering for backdoor detection. 
\textbf{FreqFed}~\citep{fereidooni2023freqfed} tackled backdoor attacks in a unique way where it transforms model updates into the frequency domain and identify unique components from malicious model updates. 
\textbf{AGSD}~\citep{ali2024adversarially} utilized a trust index history with FGSM optimization to resist backdoor attacks. Other types of defenses mitigate poisoning attacks by suppressing or perturbing model updates. Xie et al. \citep{xie2021crfl} propose to apply clip and smoothing on model parameters to control the global model smoothness, which yields robustness certification on backdoors. FLAME \citep{nguyen2022flame} adds noise to eliminate backdoors based on the concept of differential privacy. To the best of our knowledge, we are the first to transform the challenging backdoor mitigation in a non-IID scenario into an IID case.

\section{System Setting and Threat Model}

In the FL system, at the system onset, PS initializes $\Theta$. Then each training iteration works as follows: (1) PS first selects multiple clients and sends $\Theta$ to them. (2) Each of the selected clients, initializes $\mathbf{\theta}_i\!=\!\Theta$ and trains the model with its local data and provides its model update $\delta_i$ to PS. PS infers an abstract data distribution for each client, followed by overlapping clustering, feature extraction, and vote-based trust estimation. (3) PS aggregates local model updates weighted by their trust scores and updates the global model. 

We consider the data among clients to be non-IID. While the meaning of IID is generally clear, data can be non-IID in many ways. We consider differences in the data label distribution on each client, \emph{label distribution skew}, which is mostly discussed in FL\citep{awan2021contra, fang2020local, shejwalkar2021manipulating,kairouz2021advances}. 
For example, when clients are tied to particular geo-regions, the distribution of labels varies across clients — kangaroos are only in Australia or zoos; a person’s face is only in a few locations worldwide; for mobile device keyboards, certain emoji are used by one demographic but not others \citep{kairouz2021advances}.

\textbf{Threat Model: }
Attackers' primary goal is to lead the global model to classify inputs as the target labels, while simultaneously maintaining a relatively high overall accuracy to avoid being detected. Attackers have access to a dataset to train their local model. We assume they can arbitrarily manipulate their local data, which may involve injecting triggers into input data and altering labels. Additionally, they can also manipulate their model parameter updates sent to PS directly. As legitimate clients, attackers have white-box access to the global model but not to other client models. We assume PS is trusted and the defense mechanism against the poisoning attack is deployed at PS.

\section{Technical Design of \sysname}

\subsection{Challenges Identification}
In non-IID scenarios, discrepancies in model updates can arise from either inherent variations in data distributions or malicious manipulations. Determining the root cause of these model anomalies presents a formidable challenge. To tackle this problem, we propose a two-step solution. The first step involves clustering clients based on their data distribution, and the second step focuses on applying a voting-based detection methodology within each cluster. We have identified two challenges: 
\begin{itemize}
    \item \textbf{Data Distribution Inference}: With only the knowledge of model updates $\delta_i$ provided by clients, it is challenging for the defender to estimate the data distribution of clients.
    
    \item \textbf{Detection Challenge}: Traditional clustering methods cannot defend against collisions where attackers obtain the same dataset to be clustered in one cluster.  
\end{itemize}

\subsection{DDIG: Data Distribution Inference via Gradients}
\label{sec:approach-ddig}

To solve the first challenge, we propose DDIG to extract label distributions from gradients/updates.

We consider a classification task with the cross-entropy loss. Without loss of generality, we denote the gradient produced by a client in its local training process as $\nabla \theta=\frac{1}{|\mathcal{D}|}\sum_{j}\nabla_{\theta}\mathcal{L}(x_j, y_j)$, where $|\mathcal{D}|$ is the number of data samples in $\mathcal{D}$ and $x_j, y_j \in \mathcal{D}$ are sample/label pairs. 

{\theorem For each input pair $x_j, y_j$, the gradient of the logits layer $\mathbf{z} \subseteq \mathbb{R}^{m}$ (pre-softmax layer) is $\nabla_{\mathbf{z}}\mathcal{L}(x_j, y_j)= \mathbf{p}_j-\mathbf{y}_j $, where $\mathbf{p}_j \subseteq \mathbb{R}^{m}$ is a post-softmax probability vector and $\mathbf{y}_j \subseteq \mathbb{R}^{m}$ is a binary label vector with only the correct label class values 1. As all $p^{s}_{j} \in \mathbf{p}_j$ are within $[0,1]$, $\nabla_{\mathbf{z}}\mathcal{L}(x_j, y_j)$ will have a negative value only on the ``correct label class" element.}

The parameter server has no access to $\nabla_{\mathbf{z}}\mathcal{L}(x_j, y_j)$ but only the gradient. For the last linear layer $W \subseteq \mathbb{R}^{m\times r}$, where $r$ is the input feature number (i.e., the output dimension of the previous layer), the gradient of its element $W_{s, t}$ is:

 \begin{equation}
     \begin{aligned}
         \nabla W_{s, t}=\sum_{x_j, y_j} \frac{\partial \mathcal{L} (x_j, y_j)}{\partial z_s} \frac{\partial z_s}{\partial W_{s, t}}
         =\sum_{j}(p_j^s-y_j^s)o^{t}_{j}
     \end{aligned}
     \label{equ: weight gradient}
 \end{equation}
 where $o^{t}_{j} \in \mathbf{o}_j \subseteq \mathbb{R}^r$ are the inputs to the final layer, and $\mathbf{o}_j$ is non-negative when ReLU activation function is used in the penultimate layer. Therefore, for each $x_j, y_j$, $(p_j^s-y_j^s)o^{t}_{j}<0$ still holds if and only if $s$ corresponds to the correct label class. We then define an indicator ${I}_s\subseteq \mathbb{R}^m$ for each label class $(s\in[m])$:
\begin{equation}
    \begin{aligned}
        I_s=-\sum_{t=1}^{r} \nabla W_{s,t}=\sum_{j}(y_j^s-p_j^s)\sum_{t=1}^{r}{o}_{j}^t
    \end{aligned}
    \label{equ: index matrix}
\end{equation}
where $\sum_{t=1}^{r}{o}_{j}^t$ does not change with respect to $s$. Therefore, the value of $I_s$ is mainly determined by the number of samples for class $s$ in $\mathcal{D}_i$. And peaks in vector $\mathbf{I}=[I_1, I_2, ..., I_m]$ indicates more samples in the corresponding class index. 

The previous method also applies for an accumulated gradient. If we assume the clients uses the SGD optimizer to train the local model for $H$ rounds before sending the updates, the indicator vector $\mathbf{u}$ accumulates with respect to training rounds. The previous property --- the larger the elements in $\mathbf{I}$, the larger the number of samples the corresponding classes still holds.

\subsection{Data Distribution Representation}

We introduce a definition of \textbf{Abstract Data Distribution} to represent data distribution. We use $\mathcal{A}_{ij} \in \{0,1\}$ as an indicator for relative data sufficiency. $\mathcal{A}_{ij}\gets 1$ indicates that client $j$ has \emph{sufficient} \footnote{Data sufficiency is dataset specific and needs to be tuned based on the task complexity.} data for class $i$. 
A matrix $\mathcal{A}\in \{0,1\}^{m\times n}$ is used to represent the abstract data distribution information of all clients.

\begin{figure}[htbp]
    \centering
    \includegraphics[width=0.7\linewidth]{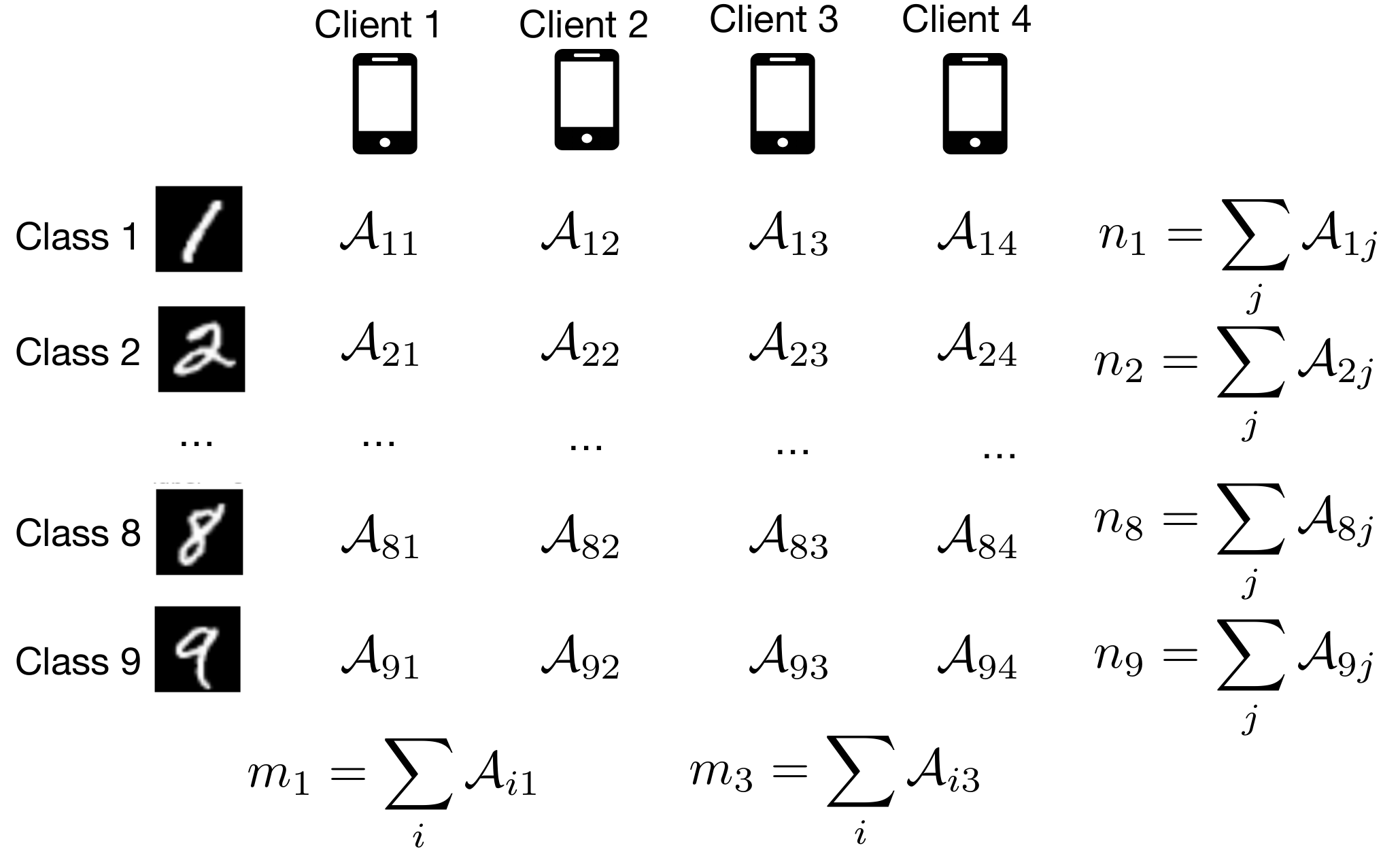}
    \caption{$\mathcal{A}_{ij}$ is an indicator of client $j$'s data sufficiency in class $i$: 1 for sufficient data and 0 for non-sufficient data. }
    \vspace{20pt}
    \label{fig:data-abstract}
\end{figure}

For each data class, a portion of clients have sufficient data due to clients' various functional environments. The number of clients having sufficient data for class $i\in[m]$ is denoted by $n_i$. Further, each client $j\in[n]$ has a subset of the $m$ classes, and their class numbers are denoted by $m_j$ as shown in Figure~\ref{fig:data-abstract}. The value satisfies $(n_i\le n), \forall i\in[m]$ and $ (m_j\le m), \forall j\in[n]$.

\begin{figure*}[htbp]
    \centering
    \includegraphics[width=0.9\linewidth]{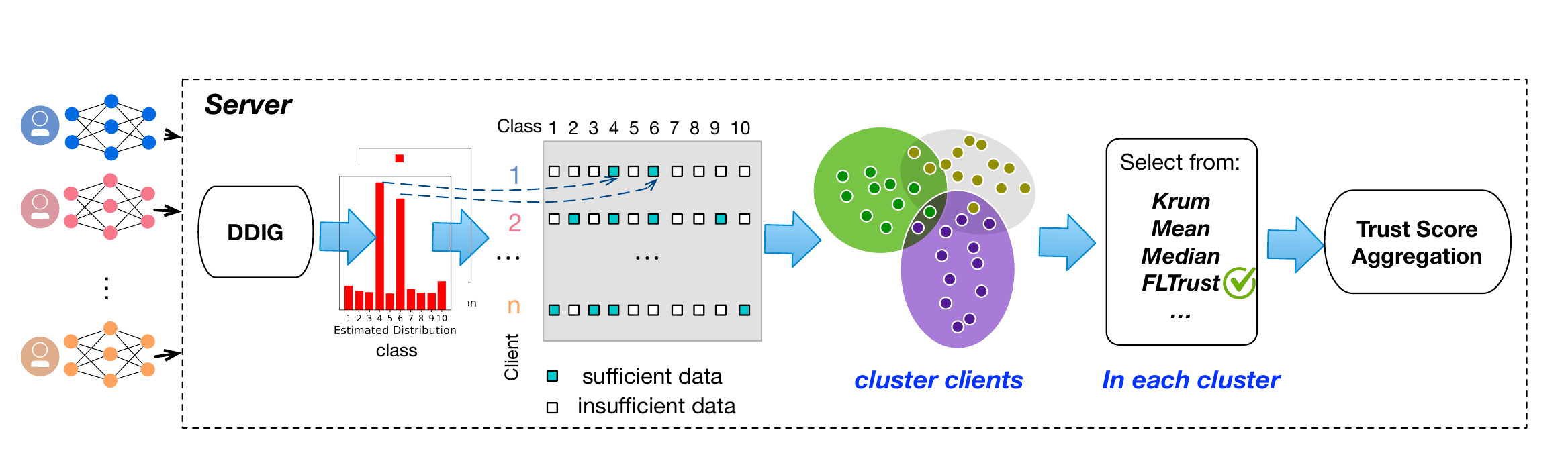}
    \vspace{-6pt}
    \caption{Overview of \sysname.}
     \vspace{12pt}
    \label{fig:workflow_server}
\end{figure*}

\subsection{Client Clustering}
\label{sec:approach-clustering}
\subsubsection{Naive Clustering}
A traditional clustering method, e.g., k-means, takes each column of $\mathcal{A}$ (e.g., one client) as input and outputs multiple groups. The results heavily depend on the selection of the number of clusters. A naive way to eliminate the dependence of an adaptively selecting a cluster number is to use the number of data classes $m$. In the naive clustering, we assign clients who possess sufficient data for data class $i\in[m]$ to cluster $i$ by default. It is worth noting that a client may belong to multiple clusters if they have sufficient data for multiple classes. But the naive clustering cannot deal with collusion, where malicious clients try to be clustered in one large group to evade detection (Neither can traditional one). To solve this problem, we introduced balanced cluster and uniform cluster inclusion.

\subsubsection{Balanced Cluster \& Uniform Cluster Inclusion} 
\label{sec:balanced_cluster}
We propose a balanced clustering scheme that features a random selection of clients in each data class, instead of including all clients with sufficient data in the cluster. The randomness can decrease the likelihood that malicious clients to form a cluster and dominate the detection. But we have not fully solved the problem. Malicious clients can collude in multiple clusters by showing sufficient data in each of the clusters. To solve this problem, we proposed a uniform cluster inclusion, meaning that, each client can participate in an identical number of clusters.

\subsubsection{Detailed Clustering Method}
\label{sec:optimal}
We use a matrix $\mathbf{x} \in \{0,1\}^{m\times n}$ to represent the clustering result, and $\mathbf{x}_{ij}$ represent client $j$ is in the cluster $i$. We set $\mathbf{x}_{ij}=1$ if client $j$ is selected to cluster $i$ and $\mathbf{x}_{ij}=0$ otherwise. We initialize the clustering as the naive cluster method --- forming a cluster with all clients having sufficient data for a specific class. In other words, the initialized $\mathbf{x} \in \{0,1\}^{m\times n}$ is equal to data sufficiency matrix $\mathcal{A}_{ij}=1$. 

To meet uniform cluster inclusion and balanced cluster, we then remove some clients from each cluster. The selection can be modeled as converting some elements with value one in matrix $\mathcal{X}$ into zeros and making no change for zero elements. The objective is to maximize the summation of elements in $x_{ij}$ (or minimize 1-$x_{ij}$) while guaranteeing uniform cluster inclusion and balanced clusters. We formulate the problem as:
\begin{equation}
\begin{aligned}
    &\text{minimize} & & \sum_{i=1}^{m}\sum_{j=1}^n (1-\mathbf{x}_{ij})*\mathcal{A}_{ij},\\
    &\text{subject to } & & \mathbf{x}_{ij} \in \{0,1\},\text{ and } x_{ij}\leq \mathcal{A}_{ij},\\
    &&&\sum_{i=1}^m \mathbf{x}_{ij} = m_{th} \mbox{, }\forall j;\text{ and }\sum_{j=1}^n \mathbf{x}_{ij}=n_{th} \mbox{, }\forall i,\\
\end{aligned}   
\end{equation}
where $m_{th}$ denotes the maximum allowed cluster size, $n_{th}$ represents the maximum number of clusters one client can participate in. The details about how the two thresholds are decided are in the Appendix. The \textbf{first condition} ensures we only select clients with sufficient data. The second line of condition corresponds to the goal of uniform cluster inclusion and balanced cluster discussed above. 
The optimization is a non-convex optimization problem and by definition is NP-hard \citep{jain2017non}. It does not have a polynomial-time closed-form solution for finding the optimal point. Therefore, we use a search-based optimization method to iteratively approximate the best solution.

Our approach begins with initializing the cluster matrix $\mathbf{x}$ as $\mathcal{A}$, as described in Algorithm~\ref{alg:algorithm}. In the initialization step, each cluster $i$ contains all clients with sufficient data for class $i$. We create balanced clusters by removing clients from each cluster.
We proceed by iterating over each cluster (i.e., each row) in the matrix $\mathcal{A}$. If a cluster contains more clients than the specified threshold $n_{th}$, we employ a removal process to balance the cluster. The removal process involves sorting the clients in the cluster based on their data class numbers in descending order (\textbf{Lines 5 to 10}). We then remove the client with the highest data class number by changing the corresponding 1 to 0 in matrix $\mathbf{x}$. Following this, we repeat a similar procedure by iterating over each column in $\mathbf{x}$ (\textbf{Lines 11 to 16}). This process of iterating and applying removal actions continues until the condition $\text{any}(RowCount) > n_{th}$ or $\text{any}(ColCount) > m_{th}$ is no longer satisfied.

It is important to note that the greedy algorithm may result in a sub-optimal solution, meaning that the result may not perfectly satisfy perfectly balanced clusters. However, in the context of backdoor attack detection, a sub-optimal solution is sufficient for achieving effective and reliable results. Thus, the proposed greedy algorithm strikes a practical balance between computational efficiency and detection accuracy, making it a valuable component of our overall approach.

\subsection{Summarized Workflow}

On the \textbf{client side}, following the traditional FL process, clients receive global model $\Theta$ from PS and use it to initialize their local model. Each client continues to train the global model using their local data and uploads model update $\delta_i$ to PS.

On the \textbf{server side}, the workflow is shown as Figure~\ref{fig:workflow_server}: 
\begin{itemize}
    \item PS first performs DDIG to infer the data distributions of each client.
    \item PS clusters the clients into overlapped clusters.
    \item In the evaluation part, we utilized client-wise cosine similarity as a detection method. Instead of obtaining a binary detection result in this step, we obtain the nearest neighbor of each client. We use $K_i$ to represent the number of times that Client $i$ is selected as the nearest neighbor of others. We can also select an existing backdoor detection method (e.g., Krum, FLDetect) and apply it to each of the clusters. 
    \item \textit{Voting-based trust estimation:} The final trust of user $i$ is calculated by $T_i=\frac{exp(K_i)}{\sum_{i=1}^n exp(K_i)}$ using its voting counts $K_i$.
    \item PS aggregated all the received model parameters weighted by their accumulated trust score, which can be represented as
$\Theta = \Theta - \lambda \sum_{i=1}^n {T_i} \frac{\delta_i}{\lVert \delta_i\rVert}$ 
where $\lambda$ represents the learning rate, and $\frac{\delta_i}{\lVert \delta_i\rVert}$ denotes the normalized gradient of client $i$. Another iteration starts.
\end{itemize}

\section{Evaluations}

We implement \sysname on the TensorFlow platform and run the experiments on a server equipped with an Intel Core i7-8700K CPU \@ 3.70GHz$\times$12, a GeForce RTX 2080 Ti GPU, and Ubuntu 18.04.3 LTS. 

\subsection{Experimental Setting}
\label{sec: setting}
\noindent\textbf{Federated learning settings: }In the studied FL system, we set the client number as $n=50$ and the per-round selection ratio as 0.2, i.e., 10 clients will be selected in each FL round. To simulate the real-world scenario, we randomize the malicious clients ratio from 0\% to 50\% of the total selected clients (with an average of 25\%) in each round. Each client manages a local model and trains the local model using an Adam optimizer with a learning rate of 0.001. A client trains its local model for five epochs before submitting the model updates. The number of total FL iterations is $T=60$. We run each experiment \emph{three} times and show the average performance.

\noindent\textbf{Datasets: }We evaluate \sysname on four datasets including three image datasets---fMNIST dataset \citep{xiao2017fashion}, CIFAR-10 dataset \citep{krizhevsky2009learning}, MNIST dataset \citep{deng2012mnist}, and one text dataset---Sentiment-140. The model architectures for the three datasets are a plain CNN, VGGNet \citep{Simonyan15}), AlxeNet \citep{krizhevsky2012imagenet}, and LSTM, respectively.

\textbf{MNIST} \citep{deng2012mnist} is a dataset of handwritten digits. It consists of 60,000 training records and 10,000 testing records, each is a $28\times28\time3$ grayscale image. \textbf{fMNIST} consists of a training set of 60,000 records and a test set of 10,000 records. Each data record is a $28\times 28$ grayscale image, associated with a label from 10 classes, including T-shirt, Trouser, Pullover, Dress, Coat, Sandal, Shirt, Sneaker, Bag, and Ankle boot.
\textbf{CIFAR-10} \citep{krizhevsky2009learning} consists of 60000 32x32 colour images in 10 classes, with 6,000 images per class. There are 50,000 training images and 10000 test images. Each image is from one of the ten classes: airplane, automobile, bird, cat, deer, dog, frog, horse, ship, and truck. \textbf{Sentiment-140} \citep{go2009twitter} consists of 1.6 million tweets, annotated with positive or negative sentiments. We adopt the same attack settings on the Sentiment-140 dataset as described in \citep{wang2020attack}. We synthetically generated the non-IID dataset following the methodology in \citep{mcmahan2017communication}. Details of this setting are shown in Sec~\ref{sec:non-iid}.
 
\noindent\textbf{Backdoor Settings:} We assume malicious clients possess clean data records and poisoned data records. Each client has a dataset of 1,200 (1,000) local training data records in MNIST and fMNIST datasets (CIFAR dataset). For each malicious client, we vary the number of backdoor data used for local training from 1 to 500 (i.e., malicious ratio 0 to around 0.3) to maintain a trade-off between stealthy and attack effectiveness.

\begin{algorithm}[tb]

\caption{A Greedy Algorithm for Clustering}
\label{alg:algorithm}
\textbf{Input}: Abstract of data distribution $\mathcal{A}\in\{0,1\}^{m \times n}$, $m$ indicates the number of classes, $n$ indicates the number of clients. \\
\textbf{Output}: \parbox[t]{\textwidth}{Cluster matrix $\mathbf{x} \in \{0, 1\}^{m \times n}$.}
\begin{algorithmic}[1] 
\STATE Set the cluster size $n_{th}=\mbox{min}(n_1, n_2, ...., n_m)$;
\STATE $m_{th}=\mbox{mean}(m_1, m_2, ..., m_n)$ \# \emph{one client can participate in $m_{th}$ clusters}.
\STATE $\mathbf{x} \gets \mathcal{A}$ \# Initialization
\WHILE{any$(RowCount) > n_{th}$ or any$(ColCount) > m_{th}$}
\FOR{$1\leq i \leq m$}
\IF{$RowCount_i> n_{th}$}
    \STATE $MaxColID \gets \mbox{argmax}(ColCounts)$
    \STATE $\mathbf{x}_{i,MaxColID} \gets 0$
\ENDIF
\ENDFOR
\FOR{$1\leq j \leq n$}
\IF{$ColCount_j> m_{th}$}
    \STATE $MaxRowID \gets \mbox{argmax}(RowCounts)$
    \STATE $\mathbf{x}_{MaxRowID,j} \gets 0$
\ENDIF
\ENDFOR
\ENDWHILE
\end{algorithmic}
\end{algorithm}

\subsection{Evaluation Metrics and Baselines}

\textbf{Main task accuracy} is evaluated on the whole test dataset, just as an attack-free scenario.  
\textbf{Attack success rate (ASR)} refers to the percentage of samples with trigger patterns that are successfully classified as the target label.


We selected six attacks including BadNet \citep{gu2017badnets}, Alternate \citep{bhagoji2019analyzing}, DBA \citep{xie2019dba}, Sybil \citep{fung2018mitigating}, Neurotoxin~\citep{zhang2022neurotoxin}, and IBA~\citep{nguyen2024iba}, reflecting a comprehensive spectrum of poisoning strategies in FL. We selected eight defenses, including Krum \citep{blanchard2017machine}, Median \citep{yin2018byzantine}, Trim \citep{yin2018byzantine}, FLTrust \citep{cao2021fltrust}, CONTRA \citep{awan2021contra}, AGSD~\citep{ali2024adversarially}, MESAS~\citep{MESAS}, and FreqFed~\citep{fereidooni2023freqfed} as our core baselines based on their widespread adoption\citep{fung2018mitigating, li2020learning, yin2018byzantine, guerraoui2018hidden}. 

\subsection{Data Distribution and Inference Accuracy }

\begin{figure}[tbp]
    \centering
    \includegraphics[width=0.7\linewidth]{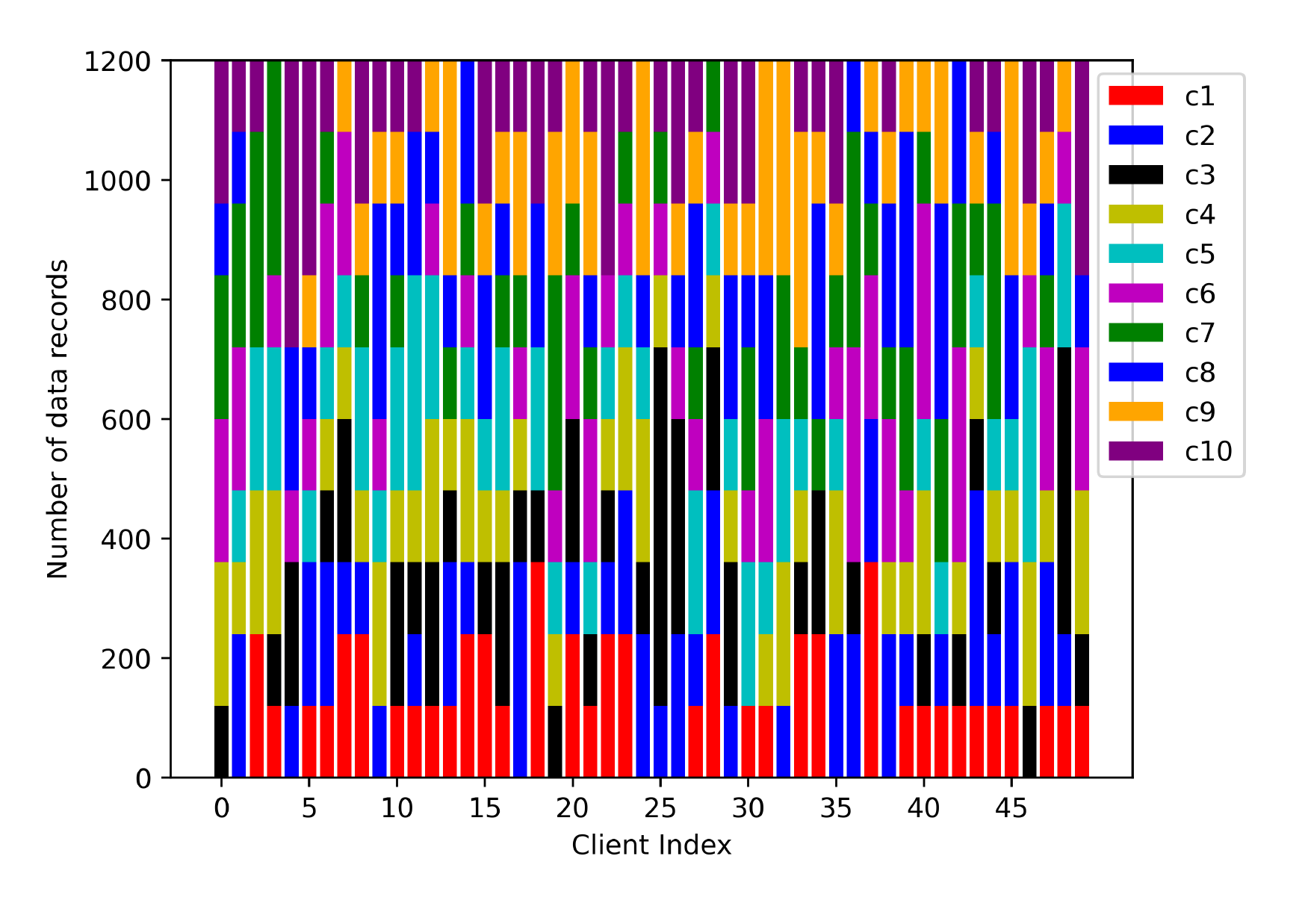}
    \vspace{-10pt}
    \caption{The illustration of data distribution. Each client has a subset of the 10 data classes (c1-c10).}
    \label{fig:cluster-illustration}
    \vspace{10pt}
\end{figure}

Figure~\ref{fig:cluster-illustration} provides a visual representation of the non-iid data distributions among clients. In the left subfigure, different colors are utilized to indicate distinct data classes, and the height of each color bar represents the corresponding data amount. As observed, clients have\textbf{ 5-7 }data classes (\textbf{out of 10}) and different data amounts per class.

\label{sec:attack-intro}

\begin{figure}[htbp]
    \centering
    \includegraphics[width=0.7\linewidth]{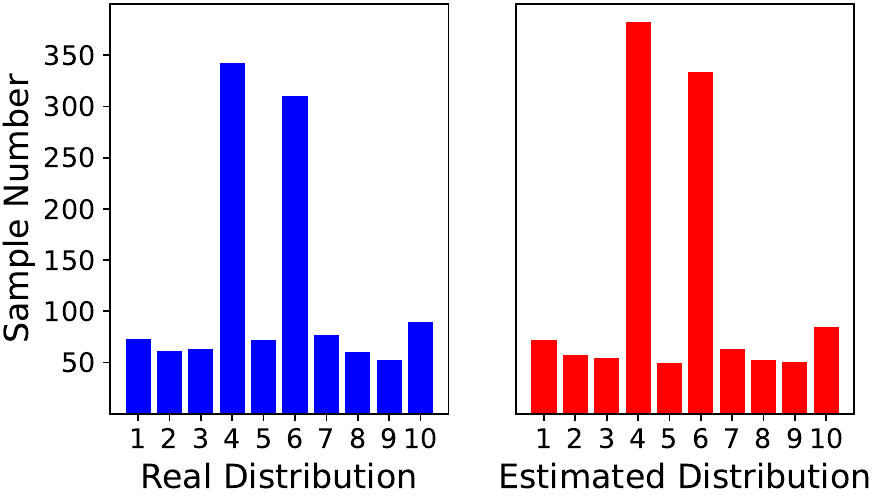}
    \vspace{-3pt}
    \caption{Data distribution inference result. }
    \vspace{20pt}
    \label{fig: label inference sample}
\end{figure}

In Figure \ref{fig: label inference sample}, we demonstrate the data distribution inference performance. We plot the distribution of the target client with 1200 samples, along with the estimated distribution from the server. The x-axis of the figures refers to different classes, and the y-axis refers to sample numbers. We can observe that the estimated data distribution is almost the same as the real one. 

\subsection{Overall Performance under Non-IID Data}
\label{sec:result-noniid}
TABLE~\ref{tab:asr} shows the ASRs of six attacks under ten defenses. From the table, we can see that \sysname outperforms other defenses by achieving the lowest ASR. We utilized heatmap color to show the differences among defenses, where green color indicates a better performance. We further present the main task accuracy of the final global model in TABLE~\ref{tab:accuracy}. We can see that \sysname achieves higher or comparable accuracy when compared with other baselines. In summary, \sysname not only reduces the ASR but also helps maintain a better model accuracy, indicating its superiority in securing FL systems in non-IID scenarios.

\begin{table*}[ht]
\caption{Backdoor \textbf{attack success rate} under defenses ($\downarrow$).}
\vspace{10pt}
\centering
\scriptsize
\setlength{\tabcolsep}{7pt}   
\renewcommand{\arraystretch}{0.8}{ 
\begin{tabular}{ccRRRRRRRRRR}
\toprule
{Dataset} & {Attack}
  & {FedAvg} & {Krum} & {Median} & {Trim} & {FLTrust}
  & {CONTRA} & {MESAS} & {FreqFed} & {AGSD} & {\sysname}\EndTableHeader\\ 
\midrule
\multirow{6}{*}{MNIST}
  & Alternate
    & 0.122 & 0.645 & 0.047 & 0.103 & 0.201
    & 0.363 & 0.700   & 0.257      & 0.172     & 0 \\
  & BadNet & 0.494 & 0     & 0.023 & 0.452 & 0.205
    & 0.343 & 0.379   & 0.317      & 0.452     & 0 \\
  & DBA & 0.020 & 0.135 & 0.002 & 0.018 & 0.021
    & 0.170 & 0.158   & 0.100      & 0.008     & 0 \\
  & Sybil & 0.117 & 0.666 & 0.048 & 0.125 & 0.232
    & 0.107 & 0.009   & 0.048      & 0.031     & 0 \\
  & Neurotoxin & 0.255 & 0.210 & 0.420 & 0.783 & 0.230
    & 0.300 & 0.057   & 0.120      & 0.115     & 0.005 \\
  & IBA & 0.206 & 0.520 & 0.100 & 0.189 & 0.202
    & 0.998 & 0.283   & 0.240      & 0.191     & 0.002 \\
\midrule
\multirow{6}{*}{fMNIST}
  & Alternate 
    & 0.667 & 0.953 & 0.786 & 0.714 & 0.915 & 0.154
    & 0.359   & 0.019      & 0.004     & 0.001 \\
  & BadNet
    & 0.828 & 0.680 & 0.811 & 0.804 & 0.836 & 0.566
    & 0.557   & 0.470      & 0.037     & 0    \\
  & DBA
    & 0.452 & 0.071 & 0.428 & 0.539 & 0.507 & 0.342
    & 0.017   & 0.032      & 0.010     & 0     \\
  & Sybil
    & 0.678 & 0.971 & 0.813 & 0.736 & 0.448 & 0.039
    & 0.026   & 0.103      & 0.053     & 0.001 \\
 & Neurotoxin
    & 0.114 & 0.250 & 0.190 & 0.319 & 0.169
    & 0.104 & 0.081   & 0.076      & 0.097     & 0.003 \\
  & IBA
    & 0.355 & 0.662 & 0.566 & 0.195 & 0.480
    & 0.212 & 0.103   & 0.110      & 0.170     & 0.001 \\
\midrule
\multirow{6}{*}{CIFAR-10}
  & Alternate
    & 0.692 & 0.994 & 0.597 & 0.734 & 0.593 & 0.249
    & 0.071   & 0.063      & 0.054     & 0.002 \\
  & BadNet
    & 0.960 & 1.00  & 0.951 & 0.802 & 0.309 & 0.379
    & 0.062   & 0.018      & 0.023     & 0.003 \\
  & DBA
    & 0.178 & 0.406 & 0.200 & 0.260 & 0.216 & 0.354
    & 0.129   & 0.083      & 0.015     & 0.009 \\
  & Sybil
    & 0.627 & 0.880 & 0.531 & 0.655 & 0.574 & 0.560
    & 0.074   & 0.094      & 0.016     & 0.004 \\
 & Neurotoxin
    & 0.981 & 0.867 & 0.130 & 0.235 & 0.182
    & 0.158 & 0.364   & 0.130      & 0.100     & 0.001 \\
  & IBA
    & 0.104 & 0.137 & 0.288 & 0.352 & 0.267
    & 0.145 & 0.0082   & 0.019      & 0.112     & 0 \\
\midrule
\multirow{6}{*}{Sentiment-140}
  & Alternate
    & 0.127 & 0.260 & 0.103 & 0.209 & 0.042 & 0.087
    & 0.106   & 0.087      & 0.051     & 0.002 \\
  & BadNet
    & 0.595 & 0.283  & 0.180 & 0.077 & 0.062 & 0.079
    & 0.034   & 0.153      & 0.012     & 0.001 \\
  & DBA
    & 0.965 & 0.710 & 0.763 & 0.751 & 0.265 & 0.271
    & 0.197   & 0.072      & 0.080     & 0.003 \\
  & Sybil
    & 0.210 & 0.653 & 0.305 & 0.655 & 0.259 & 0.272
    & 0.073   & 0.090      & 0.057     & 0.001 \\
  & Neurotoxin
    & 0.979 & 0.846 & 0.335 & 0.300 & 0.185
    & 0.098 & 0.107   & 0.100      & 0.066     & 0.007 \\
  & IBA
    & 0.533 & 0.347 & 0.521 & 0.136 & 0.143
    & 0.109 & 0.121   & 0.020      & 0.045     & 0.003 \\
\bottomrule
\multicolumn{12}{l}{Heatmap is used to indicate performance level, where green color represents better performance, yellow to red represents worse performance.}
\end{tabular}
}
\label{tab:asr}
\end{table*}

\begin{table*}[ht]
\caption{Main task \textbf{accuracy} ($\uparrow$). }
\scriptsize
\vspace{8pt}
\setlength{\tabcolsep}{8pt}   
\renewcommand{\arraystretch}{0.8}{ 
\vspace{-8pt}
\begin{center}
\begin{tabular}{cccccccccccc}
\toprule
{Dataset} & {Attack}
& {FedAvg} & {Krum} & {Median} & {Trim} & {FLTrust}
  & {CONTRA} & {MESAS} & {FreqFed} & {AGSD} & {\sysname} \EndTableHeader\\ 
\midrule
\multirow{6}{*}{MNIST} 
&Attack-free & 0.992  & 0.991 & 0.992 & 0.992 & 0.992 & 0.990 & 0.991 & 0.990  & \textbf{0.993} & 0.991 \\
&Alternate & 0.991 & 0.991 & 0.990 & 0.991 & 0.991 & 0.989   & \textbf{0.992}   & 0.983  & 0.990 & 0.991 \\
&BadNet& 0.991 & 0.991 & 0.990 & 0.991  & 0.989 & 0.988   & 0.985   & \textbf{0.992}  & 0.990 & 0.991\\
&DBA & \textbf{0.992} & 0.989 & \textbf{0.992} & \textbf{0.992} & 0.991 & 0.866  & 0.991   & 0.989    & 0.991  & 0.990\\
&Sybil & \textbf{0.991} & 0.990 & \textbf{0.991} & \textbf{0.991} & \textbf{0.991} & 0.987  & 0.985   & \textbf{0.991}   & 0.990  & \textbf{0.991}\\
&Neurotoxin& 0.990 & 0.989 & 0.991 & 0.990 & 0.988 & 0.990 & 0.992  & 0.991  & \textbf{0.993}     & 0.992\\
&IBA & 0.986 & 0.988 & 0.987 & 0.989 & 0.985 & 0.990 & 0.991 & \textbf{0.992} & 0.990 & 0.991 \\
\midrule
\multirow{5}{*}{fMNIST} 
&Attack-free& \textbf{0.911}  & 0.901 & 0.909  & 0.910 & 0.905   & 0.890 & 0.899 & 0.897 & 0.900  & 0.902\\
&Alternate & \textbf{0.908} & 0.895 & 0.904 & 0.906 & 0.897 & 0.886  & 0.893 & 0.898  & 0.890  &  0.897 \\
&BadNet & \textbf{0.903} & 0.610 & 0.893 & \textbf{0.903} & 0.887 & 0.612  & 0.891   & 0.899  & 0.893  & \textbf{0.903} \\
&DBA& 0.894 & 0.881 & 0.897 & 0.895 & 0.876 & 0.613  & 0.889  & 0.891   & 0.887  & \textbf{0.900} \\
&Sybil & \textbf{0.907} & 0.895 & 0.904 & 0.906 & 0.906 & 0.889  & 0.894 & 0.898  & 0.895  & 0.903 \\
&Neurotoxin & 0.901 & 0.894 & 0.899 & 0.902 & 0.896 & 0.885 & 0.895  & 0.892  & 0.896  & \textbf{0.904}\\
&IBA& 0.896 & 0.890 & 0.893 & 0.895 & 0.887 & 0.889 & 0.891 & 0.894  & 0.890  & \textbf{0.900}\\
\midrule
\multirow{5}{*}{CIFAR-10} 
& Attack-free  & 0.711  & 0.691 & 0.709  & 0.702 & 0.721 & \textbf{0.722}  & 0.698 & 0.710 & 0.703 & 0.720\\
& Alternate & 0.687 & 0.653 & 0.681 & 0.681 & 0.705 & 0.676  & 0.682 & 0.688  & 0.691  & \textbf{0.709}\\
& BadNet & 0.683 & 0.662 & 0.675 & 0.679 & 0.495 & 0.586  & 0.683 & 0.690 & 0.695  & \textbf{0.716}\\
& DBA & 0.677 & 0.638 & 0.668 & 0.667 & 0.675 & 0.466  & 0.673  & 0.681  & 0.687  & \textbf{0.703}\\
& Sybil & 0.693 & 0.668 & 0.679 & 0.685 & 0.681 & 0.592 & 0.686  & 0.689  & 0.696  & \textbf{0.717} \\
& Neurotoxin & 0.690 & 0.665 & 0.674 & 0.678 & 0.672 & 0.650 & 0.679   & 0.684  & 0.688 & \textbf{0.698} \\
& IBA & 0.672 & 0.661 & 0.670 & 0.675 & 0.668 & 0.659 & 0.676  & 0.681  & 0.685  & \textbf{0.693} \\
\midrule
\multirow{5}{*}{Sentiment-140} 
&Attack-free  & 0.768 & 0.759 & 0.763 & 0.770 & 0.752 & 0.760 & 0.765 & 0.769 & 0.772 & \textbf{0.781} \\
&Alternate & 0.782 & 0.774 & 0.779 & \textbf{0.783} & 0.770 & 0.768  & 0.775 & 0.780  & 0.781  & 0.781 \\
&BadNet& 0.776 & 0.763 & 0.772 & 0.771 & 0.765 & 0.752  & 0.770 & 0.773  & 0.778  &\textbf{0.785} \\
&DBA & 0.765 & 0.754 & 0.760 & 0.759 & 0.750 & 0.742  & 0.758 & 0.765 & 0.769  & \textbf{0.780}\\
&Sybil & 0.780 & 0.768 & 0.775 & 0.777 & 0.774 & 0.760 & 0.773 & 0.776  & 0.779 & \textbf{0.786}\\
&Neurotoxin& 0.762 & 0.745 & 0.751 & 0.757 & 0.746& \textbf{0.780} & 0.750   & 0.776 & 0.763 & 0.779 \\
& IBA& 0.749 & 0.735 & 0.742 & 0.748 & 0.739& 0.732 & 0.745 & 0.751 & 0.755 & \textbf{0.775} \\
\bottomrule

\end{tabular}
\label{tab:accuracy}
\end{center}
}
\end{table*}

\begin{figure}[ht]
\centering
\subfigure[fMNIST]{
\begin{minipage}[t]{0.45\linewidth}
\label{fig:box-fmnist}
\centering
\includegraphics[width=3cm]{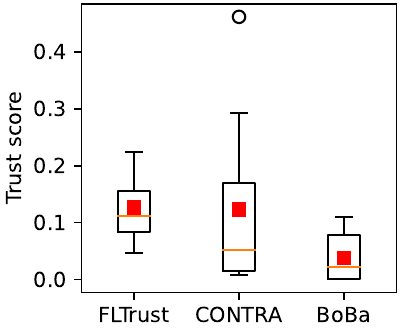}
\end{minipage}%
}
\subfigure[CIFAR-10]{
\begin{minipage}[t]{0.45\linewidth}
\label{fig:box-cifar}
\centering
\includegraphics[width=3cm]{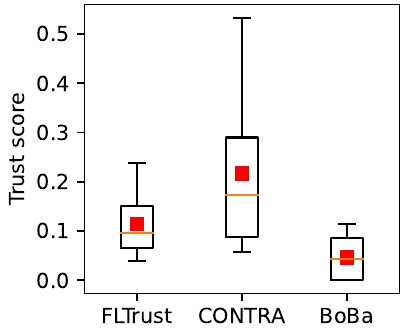}
\end{minipage}
}
\centering
\vspace{-4pt}
\caption{Trust scores of \textbf{MALICIOUS} clients (the lower the better). Orange lines represent the median and red squares represent the mean. }
\vspace{5pt}
\label{fig:mal-trust}
\end{figure}

To further illustrate the detection performance, we present the trust score distributions of malicious clients in Figure~\ref{fig:mal-trust}. We scale the trust scores for all selected clients to make sure they sum up to one. For decrease the impact of backdoor attacks, ideally, we want to reduce the trust score for malicious clients as close as possible to zero. As we can see from Figure~\ref{fig:mal-trust}, \sysname achieves the lowest average trust scores compared with the other two trust-based defenses---FLTrust and CONTRA, indicating that \sysname has a better detection performance against backdoor attacks.

\subsection{Performance under Adaptive Attack}

\begin{figure}[htbp]
    \centering
    \includegraphics[width=0.5\linewidth]{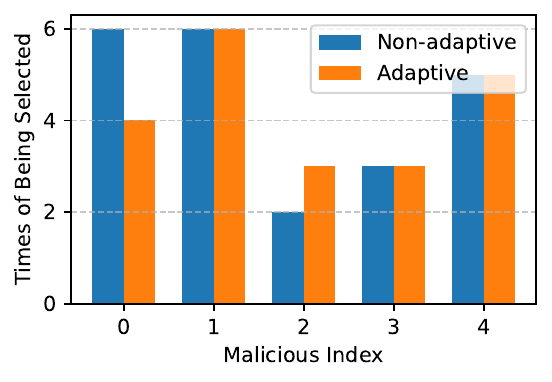}
    \caption{Times of malicious client being selected under adaptive attack.  }
    \vspace{5pt}
    \label{fig:adaptive}
\end{figure}

An adaptive attack refers to an attack that is specifically designed to defeat the proposed defense based on knowledge of the defense strategy. We initialize an adaptive attack against the proposed clustering module. We assume that an adaptive attacker can successfully mislead DDIG by manipulating their gradients. For collusion purposes, they may present data covering all categories to increase their chances of being included in clusters. For better illustrate the effectiveness of adaptive attacks, we record the number of times malicious clients are selected throughout the learning rounds. From Figure~\ref{fig:adaptive}, we observe that, compared to the base attack (Alternate attack), the adaptive attacker does not increase their chance of being selected. Besides the selected times, we also evaluated the final ASR of the adaptive attack and base attack. Adaptive attack achieves 0.0004 ASR, which shows no obvious increase compared to the non-adaptive attack (ASR = 0.00015). This resilience is due to uniform cluster inclusion, which ensures that each client participates in the same—or a comparable—number of clusters, regardless of their data sufficiency. We acknowledge that there are many other possible adaptive attack strategies, that can be explored in future research.

\subsection{Robustness against Various Settings}

\begin{figure}[ht]
\centering
\includegraphics[width=0.9\linewidth]{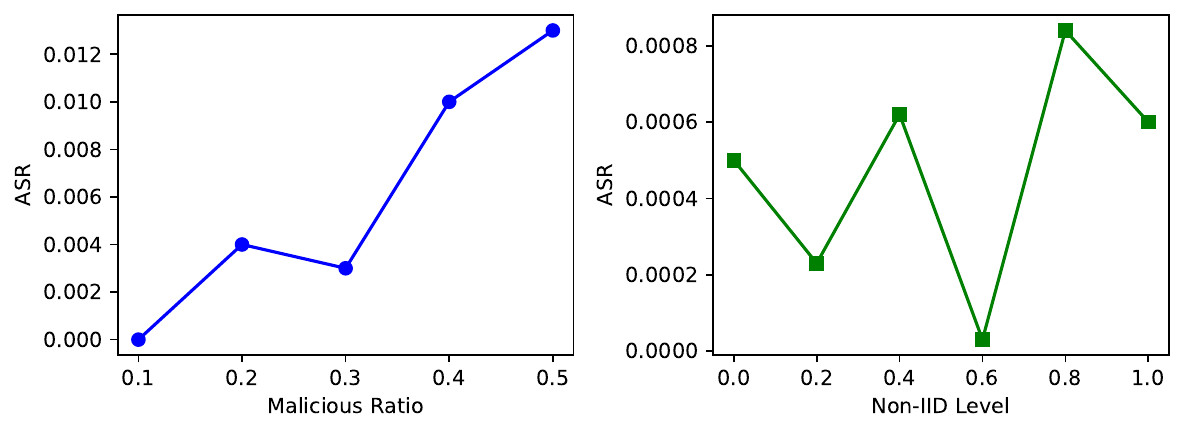}
\label{fig:mal-fraction}
\vspace{-5pt}
\caption{ASR under various malicious client ratios and non-IID degrees.}
\vspace{8pt}
\label{fig:factors}
\end{figure}

We first evaluate the performance of \sysname under various malicious client fractions. As shown in Figure~\ref{fig:factors}, the system maintains a high detection performance when the malicious client number is less than 50\%. We see the performance degrades promptly when the malicious ratio is larger than 50\% which is not shown in this figure. We note that we do not focus on the part where the malicious ratio is larger than 50\% as we assume benign clients are the majority. 

We evaluate the performance under different non-IID degree and plot the results in Figure~\ref{fig:factors}. As we can see from the figure, ASR is not sensitive to non-IID degree. Here, a value from 0 to 1 is used to indicate the non-IID degree, where 0 indicates IID data and 1 indicates fully non-IID data. Suppose we use $p$ to indicate the non-IID degree, and the total number of data records is $N$. In experiments, we first assign the $(1-p)$ portion of the dataset uniformly to all clients (i.e., IID data). In the second step, we sort the remaining $p*N$ data by digit label, divide it into 1000 shards, and assign each of the 50 clients 20 shards. This is a pathological non-IID partition of the data. We can see that \sysname also achieves a low ASR in the IID data setting and a fully non-IID scenario. 

\label{sec:non-iid}

\subsection{Computation Overhead}
\label{overhead}

Compared to traditional FL systems, the server has two additional tasks in \sysname---DDIG, client clustering. DDIG is a deterministic method by directly summing the last layer gradients as shownin Equation~\ref{equ: index matrix}. The time complexity is negligible. The time complexity of the proposed greedy algorithm for clustering is shown in Figure~\ref{tab:runtime}. We further report the full end-to-end runtime (including DDIG computation, clustering, voting, and aggregation) in comparison to standard aggregation-based defenses on the MNIST dataset with 200 clients under the BadNet attack. Across 50 rounds, BoBa achieves an average runtime of 0.0523 seconds/round, which is approximately twice that of FedAvg (no defense) with a runtime of 0.0231 seconds/round.

\begin{table}[h]
\caption{Computation Time of Clustering.}
\vspace{8pt}
    \centering
    \begin{tabular}{c|c|c|c|c|c|c|c}
    \toprule
        Client Num & 10 & 20 & 30 & 40 & 50 & 100 & 200  \\
        Runtime (ms) &  0.83 & 1.21 & 1.57 & 1.87 & 2.20 & 2.62 & 3.75 \\
        \bottomrule
    \end{tabular}
    \label{tab:runtime}
\end{table}

\subsection{Discussions}
BoBa performs best when some clients have distinct label distributions, as this supports effective clustering. In cases where client distributions are overlapping, DDIG’s signal will weaken, but in such cases, the method will work as in IID scenarios, where BoBa has been shown to generalize well(Figure~\ref{fig:factors}). 
If malicious clients attempt to mimic diverse distributions, the defense may degrade. Nonetheless, uniform cluster inclusion ensures that even boundary clients do not exert disproportionate influence across multiple clusters. 

\section{Conclusions}
We proposed \sysname, an effective solution for mitigating backdoor attacks in FL with non-IID data. By introducing overlapping clustering with the DDIG module, we establish \sysname as a powerful performance booster for detection mechanisms against various backdoor attack scenarios. We propose two constraints --- Balanced Clustering and Uniform Inclusion in the clustering algorithm to solve the collusion problem. Our extensive evaluation showcases the superiority of \sysname over other baseline methods, as it consistently achieves lower ASRs across various attack strategies and non-IID levels on multiple datasets. Moreover, the extensive evaluation results confirm the robustness of \sysname, positioning it as a promising safeguard for FL applications.

\begin{ack}
This work was supported in part by the Office of Naval Research under grants N00014-24-1-2730, the US National Science Foundation under grants 2154929, 2247560, 2247561, 2235232, and 2312447.
\end{ack}


\bibliography{ref}

@inproceedings{ali2024adversarially,
  title={Adversarially guided stateful defense against backdoor attacks in federated deep learning},
  author={Ali, Hassan and Nepal, Surya and Kanhere, Salil S and Jha, Sanjay},
  booktitle={2024 Annual Computer Security Applications Conference (ACSAC)},
  pages={794--809},
  year={2024},
  organization={IEEE}
}

@inproceedings{awan2021contra,
  title={Contra: Defending against poisoning attacks in federated learning},
  author={Awan, Sana and Luo, Bo and Li, Fengjun},
  booktitle={26th European Symposium on Research in Computer Security},
  pages={455--475},
  year={2021},
  organization={Springer}
}

@inproceedings{bagdasaryan2020backdoor,
  title={How to backdoor federated learning},
  author={Bagdasaryan, Eugene and Veit, Andreas and Hua, Yiqing and Estrin, Deborah and Shmatikov, Vitaly},
  booktitle={International Conference on Artificial Intelligence and Statistics},
  pages={2938--2948},
  year={2020},
  organization={PMLR}
}

@inproceedings{bhagoji2019analyzing,
  title={Analyzing federated learning through an adversarial lens},
  author={Bhagoji, Arjun Nitin and Chakraborty, Supriyo and Mittal, Prateek and Calo, Seraphin},
  booktitle={International Conference on Machine Learning},
  pages={634--643},
  year={2019},
  organization={PMLR}
}

@inproceedings{blanchard2017machine,
  title={Machine learning with adversaries: Byzantine tolerant gradient descent},
  author={Blanchard, Peva and Guerraoui, Rachid and Stainer, Julien and others},
  booktitle={NeurIPS},
  year={2017}
}

@inproceedings{briggs2020federated,
  title={Federated learning with hierarchical clustering of local updates to improve training on non-IID data},
  author={Briggs, Christopher and Fan, Zhong and Andras, Peter},
  booktitle={2020 International Joint Conference on Neural Networks (IJCNN)},
  pages={1--9},
  year={2020},
  organization={IEEE}
}

@inproceedings{cao2021fltrust,
  title={FLTrust: Byzantine-robust Federated Learning via Trust Bootstrapping},
  author={Cao, Xiaoyu and Fang, Minghong and Liu, Jia and Gong, Neil Zhenqiang},
  booktitle={Network and Distributed Systems Security Symposium NDSS},
  year={2021}
}

@article{chen2017distributed,
  title={Distributed statistical machine learning in adversarial settings: Byzantine gradient descent},
  author={Chen, Yudong and Su, Lili and Xu, Jiaming},
  journal={Proceedings of the ACM on Measurement and Analysis of Computing Systems},
  volume={1},
  number={2},
  pages={1--25},
  year={2017},
  publisher={ACM New York, NY, USA}
}

@article{deng2012mnist,
  title={The mnist database of handwritten digit images for machine learning research},
  author={Deng, Li},
  journal={IEEE Signal Processing Magazine},
  volume={29},
  number={6},
  pages={141--142},
  year={2012},
  publisher={IEEE}
}

@inproceedings{fang2020local,
  title={Local model poisoning attacks to Byzantine-robust federated learning},
  author={Fang, Minghong and Cao, Xiaoyu and Jia, Jinyuan and Gong, Neil},
  booktitle={29th $\{$USENIX$\}$ Security Symposium},
  pages={1605--1622},
  year={2020}
}

@inproceedings{fereidooni2023freqfed,
  title={FreqFed: A Frequency Analysis-Based Approach for Mitigating Poisoning Attacks in Federated Learning},
  author={Fereidooni, Hossein and Pegoraro, Alessandro and Rieger, Phillip and Dmitrienko, Alexandra and Sadeghi, Ahmad-Reza},
  booktitle={Network and Distributed Systems Security Symposium NDSS},
  year={2024}
}

@article{fung2018mitigating,
  title={Mitigating sybils in federated learning poisoning},
  author={Fung, Clement and Yoon, Chris JM and Beschastnikh, Ivan},
  journal={arXiv preprint arXiv:1808.04866},
  year={2018}
}

@article{ghosh2019robust,
  title={Robust federated learning in a heterogeneous environment},
  author={Ghosh, Avishek and Hong, Justin and Yin, Dong and Ramchandran, Kannan},
  journal={arXiv preprint arXiv:1906.06629},
  year={2019}
}

@misc{go2009twitter,
  title={Twitter sentiment classification using distant supervision},
  author={Go, Alec and Bhayani, Richa and Huang, Lei},
  year={2009},
  note={CS224N Project Report, Stanford},
  url={http://help.sentiment140.com/}
}

@article{gu2017badnets,
  title={Badnets: Identifying vulnerabilities in the machine learning model supply chain},
  author={Gu, Tianyu and Dolan-Gavitt, Brendan and Garg, Siddharth},
  journal={arXiv preprint arXiv:1708.06733},
  year={2017}
}

@inproceedings{guerraoui2018hidden,
  title={The hidden vulnerability of distributed learning in byzantium},
  author={Guerraoui, Rachid and Rouault, S{\'e}bastien and others},
  booktitle={International Conference on Machine Learning},
  pages={3521--3530},
  year={2018},
  organization={PMLR}
}

@article{hard2018federated,
  title={Federated learning for mobile keyboard prediction},
  author={Hard, Andrew and Rao, Kanishka and Mathews, Rajiv and Ramaswamy, Swaroop and Beaufays, Fran{\c{c}}oise and Augenstein, Sean and Eichner, Hubert and Kiddon, Chlo{\'e} and Ramage, Daniel},
  journal={arXiv preprint arXiv:1811.03604},
  year={2018}
}

@article{jain2017non,
  title={Non-convex optimization for machine learning},
  author={Jain, Prateek and Kar, Purushottam and others},
  journal={Foundations and Trends{\textregistered} in Machine Learning},
  volume={10},
  number={3-4},
  pages={142--363},
  year={2017},
  publisher={Now Publishers, Inc.}
}

@article{kairouz2021advances,
  title={Advances and open problems in federated learning},
  author={Kairouz, Peter and McMahan, H Brendan and Avent, Brendan and Bellet, Aur{\'e}lien and Bennis, Mehdi and Bhagoji, Arjun Nitin and Bonawitz, Kallista and Charles, Zachary and Cormode, Graham and Cummings, Rachel and others},
  journal={Foundations and trends{\textregistered} in machine learning},
  volume={14},
  number={1--2},
  pages={1--210},
  year={2021},
  publisher={Now Publishers, Inc.}
}

@article{konevcny2016federated1,
  title={Federated learning: Strategies for improving communication efficiency},
  author={Kone{\v{c}}n{\`y}, Jakub and McMahan, H Brendan and Yu, Felix X and Richt{\'a}rik, Peter and Suresh, Ananda Theertha and Bacon, Dave},
  journal={arXiv preprint arXiv:1610.05492},
  year={2016}
}

@inproceedings{MESAS,
author = {Krau\ss{}, Torsten and Dmitrienko, Alexandra},
title = {MESAS: Poisoning Defense for Federated Learning Resilient against Adaptive Attackers},
year = {2023},
booktitle = {Proceedings of the 2023 ACM SIGSAC Conference on Computer and Communications Security},
pages = {1526–1540},
series = {CCS '23}
}

@techreport{krizhevsky2009learning,
  title={Learning multiple layers of features from tiny images},
  author={Krizhevsky, Alex and Hinton, Geoffrey},
  year={2009},
  institution={Citeseer}
}

@article{krizhevsky2012imagenet,
  title={Imagenet classification with deep convolutional neural networks},
  author={Krizhevsky, Alex and Sutskever, Ilya and Hinton, Geoffrey E},
  journal={Advances in neural information processing systems},
  volume={25},
  year={2012}
}

@article{li2020learning,
  title={Learning to Detect Malicious Clients for Robust Federated Learning},
  author={Li, Suyi and Cheng, Yong and Wang, Wei and Liu, Yang and Chen, Tianjian},
  journal={arXiv preprint arXiv:2002.00211},
  year={2020}
}

@inproceedings{nguyen2022flame,
  title={FLAME: Taming backdoors in federated learning},
  author={Nguyen, Thien Duc and Rieger, Phillip and De Viti, Roberta and Chen, Huili and Brandenburg, Bj{\"o}rn B and Yalame, Hossein and M{\"o}llering, Helen and Fereidooni, Hossein and Marchal, Samuel and Miettinen, Markus and others},
  booktitle={31st USENIX Security Symposium},
  pages={1415--1432},
  year={2022}
}

@article{nguyen2024iba,
  title={Iba: Towards irreversible backdoor attacks in federated learning},
  author={Nguyen, Thuy Dung and Nguyen, Tuan A and Tran, Anh and Doan, Khoa D and Wong, Kok-Seng},
  journal={in NeurIPS},
  year={2024}
}

@inproceedings{rieger2022deepsight,
  title={Deepsight: Mitigating backdoor attacks in federated learning through deep model inspection},
  author={Rieger, Phillip and Nguyen, Thien Duc and Miettinen, Markus and Sadeghi, Ahmad-Reza},
  booktitle={Network and Distributed Systems Security Symposium},
  year={2022},
}

@inproceedings{rieger2024crowdguard,
  title={Crowdguard: Federated backdoor detection in federated learning},
  author={Rieger, Phillip and Krau{\ss}, Torsten and Miettinen, Markus and Dmitrienko, Alexandra and Sadeghi, Ahmad-Reza},
  booktitle={Network and Distributed Systems Security Symposium NDSS},
  year={2024}
}

@inproceedings{shejwalkar2021manipulating,
  title={Manipulating the byzantine: Optimizing model poisoning attacks and defenses for federated learning},
  author={Shejwalkar, Virat and Houmansadr, Amir},
  booktitle={NDSS},
  year={2021}
}

@inproceedings{shen2016auror,
  title={Auror: Defending against poisoning attacks in collaborative deep learning systems},
  author={Shen, Shiqi and Tople, Shruti and Saxena, Prateek},
  booktitle={Proceedings of the 32nd Annual Conference on Computer Security Applications},
  pages={508--519},
  year={2016}
}

@InProceedings{Simonyan15,
  author       = "Karen Simonyan and Andrew Zisserman",
  title        = "Very Deep Convolutional Networks for Large-Scale Image Recognition",
  booktitle    = "ICLR",
  year         = "2015",
}

@article{so2020byzantine,
  title={Byzantine-resilient secure federated learning},
  author={So, Jinhyun and G{\"u}ler, Ba{\c{s}}ak and Avestimehr, A Salman},
  journal={IEEE Journal on Selected Areas in Communications},
  year={2020},
  publisher={IEEE}
}

@article{sun2019can,
  title={Can you really backdoor federated learning?},
  author={Sun, Ziteng and Kairouz, Peter and Suresh, Ananda Theertha and McMahan, H Brendan},
  journal={arXiv preprint arXiv:1911.07963},
  year={2019}
}

@article{wang2020attack,
  title={Attack of the tails: Yes, you really can backdoor federated learning},
  author={Wang, Hongyi and Sreenivasan, Kartik and Rajput, Shashank and Vishwakarma, Harit and Agarwal, Saurabh and Sohn, Jy-yong and Lee, Kangwook and Papailiopoulos, Dimitris},
  journal={Advances in Neural Information Processing Systems},
  volume={33},
  pages={16070--16084},
  year={2020}
}

@inproceedings{wang2022flare,
  title={Flare: defending federated learning against model poisoning attacks via latent space representations},
  author={Wang, Ning and Xiao, Yang and Chen, Yimin and Hu, Yang and Lou, Wenjing and Hou, Y Thomas},
  booktitle={Proceedings of the 2022 ACM on Asia Conference on Computer and Communications Security},
  pages={946--958},
  year={2022}
}

@article{wang2019adaptive,
  title={Adaptive federated learning in resource constrained edge computing systems},
  author={Wang, Shiqiang and Tuor, Tiffany and Salonidis, Theodoros and Leung, Kin K and Makaya, Christian and He, Ting and Chan, Kevin},
  journal={IEEE Journal on Selected Areas in Communications},
  volume={37},
  number={6},
  pages={1205--1221},
  year={2019},
  publisher={IEEE}
}

@article{xiao2017fashion,
  title={Fashion-mnist: a novel image dataset for benchmarking machine learning algorithms},
  author={Xiao, Han and Rasul, Kashif and Vollgraf, Roland},
  journal={arXiv preprint arXiv:1708.07747},
  year={2017}
}

@inproceedings{xie2019dba,
  title={Dba: Distributed backdoor attacks against federated learning},
  author={Xie, Chulin and Huang, Keli and Chen, Pin-Yu and Li, Bo},
  booktitle={ICLR},
  year={2019}
}

@inproceedings{xie2021crfl,
  title={Crfl: Certifiably robust federated learning against backdoor attacks},
  author={Xie, Chulin and Chen, Minghao and Chen, Pin-Yu and Li, Bo},
  booktitle={ICML},
  pages={11372--11382},
  year={2021},
  organization={PMLR}
}

@inproceedings{yin2018byzantine,
  title={Byzantine-robust distributed learning: Towards optimal statistical rates},
  author={Yin, Dong and Chen, Yudong and Kannan, Ramchandran and Bartlett, Peter},
  booktitle={International conference on machine learning},
  pages={5650--5659},
  year={2018},
  organization={Pmlr}
}

@inproceedings{zhang2022neurotoxin,
  title={Neurotoxin: Durable backdoors in federated learning},
  author={Zhang, Zhengming and Panda, Ashwinee and Song, Linyue and Yang, Yaoqing and Mahoney, Michael and Mittal, Prateek and Kannan, Ramchandran and Gonzalez, Joseph},
  booktitle={ICML},
  year={2022}
}

@inproceedings{zhao2019pdgan,
  title={PDGAN: A novel poisoning defense method in federated learning using generative adversarial network},
  author={Zhao, Ying and Chen, Junjun and Zhang, Jiale and Wu, Di and Teng, Jian and Yu, Shui},
  booktitle={International Conference on Algorithms and Architectures for Parallel Processing},
  pages={595--609},
  year={2019},
  organization={Springer}
}

@inproceedings{mcmahan2017communication,
  title={Communication-Efficient Learning of Deep Networks from Decentralized Data},
  author={McMahan, Brendan and Moore, Eider and Ramage, Daniel and Hampson, Seth and y Arcas, Blaise Aguera},
  booktitle={Artificial Intelligence and Statistics},
  pages={1273--1282},
  year={2017}
}

\end{document}